\documentclass[11pt]{article}

\usepackage[preprint]{acl}

\usepackage{times}
\usepackage{latexsym}

\usepackage[T1]{fontenc}
\usepackage[utf8]{inputenc}

\usepackage{microtype}

\usepackage{inconsolata}

\usepackage{graphicx}
\usepackage{booktabs}
\usepackage{multirow}
\usepackage{multicol}
\usepackage{multirow}
\usepackage{booktabs}
\usepackage{array}
\usepackage{ulem}
\usepackage{tabularx}
\usepackage{makecell}
\usepackage{amsmath}
\usepackage{amsfonts}
\usepackage{algorithm}
\usepackage{algpseudocode}
\usepackage{booktabs}
\usepackage{wrapfig}
\title{Tokenization, Fusion and Decoupling: Bridging the  Granularity Mismatch Between Large Language Models and Knowledge Graphs }

\author{Siyue Su \and Jian Yang \and Bo Li \and Guanglin Niu \\
        Beihang University }

\begin{document}
\maketitle
\begin{abstract}
Leveraging Large Language Models (LLMs) for Knowledge Graph Completion (KGC) is promising but hindered by a fundamental granularity mismatch. LLMs operate on fragmented token sequences, whereas entities are the fundamental units in knowledge graphs (KGs) scenarios. Existing approaches typically constrain predictions to limited candidate sets or align entities with the LLM's vocabulary by pooling multiple tokens or decomposing entities into fixed-length token sequences, which fail to capture both the semantic meaning of the text and the structural integrity of the graph. To address this, we propose \textbf{KGT}, a novel framework that uses dedicated entity tokens to enable efficient, full-space prediction. Specifically, we first introduce specialized tokenization to construct feature representations at the level of dedicated entity tokens. We then fuse pre-trained structural and textual features into these unified embeddings via a relation-guided gating mechanism, avoiding training from scratch. Finally, we implement decoupled prediction by leveraging independent heads to separate and combine semantic and structural reasoning. Experimental results show that KGT consistently outperforms state-of-the-art methods across multiple benchmarks.

\end{abstract}
% 整体思路：
% Llm在nlp领域表现优异但是在kg领域的探索仍处于前沿研究阶段，一个原因是llm和kg天然的不匹配性。一种自然的思路是为图元素创造新的token，但是从头学习大量新token效率极低且容易损伤llm的原生性能。
% 为此，现在主要有两类型方法：
% 第一种类型(代表方法：Filter-then-Generate: Large Language Models with Structure-Text Adapter for Knowledge Graph Completion，Self-supervised Quantized Representation for Seamlessly Integrating Knowledge Graphs with Large Language Models，KICGPT: Large Language Model with Knowledge in Context for  Knowledge Graph Completion)是向这类型困境妥协，选择了简化问题，即仅在有限候选上进行预测，依赖外部模块如传统的kge方法进行过滤，这类型方法较为低效且损失了llm的全空间预测能力
% 另外一类型基于llm原生词表组合构成实体的方法（代表方法：K-ON: Stacking Knowledge On the Head Layer of Large Language Model,MKGL: Mastery of a Three-Word Language）直接预测实体实现全空间预测，但是难以实现细粒度的文本信息和粗粒度的结构信息的平衡，k-on缺乏显式的结构先验，类似于传统的基于文本的方法，仅在loss中引入实体级别的对比；mkgl将实体文本序列压缩为一个向量，不可避免的存在语义损失。
% 为此，我们尝试从新的角度探索，提出了一个新的模型：为关系和实体创造新的token，继承预训练的原始文本和结构特征，并采用双流结构在微调中融合解耦实现平衡学习。这样解决了粒度错配问题，实现了全空间预测(对应第一类方法)，又在LLM微调中实现了文本信息和结构信息的平衡（对应第二类方法）。具体来说，首先为了避免从头学习，我们定制了kg适用的token特征提取模型：分别使用预训练的句子特征提取器和tucker模型获取文本和结构信息，并投影到llm空间，经自适应的fusion得到kg元素的的特征。其次，我们针对kg任务设计了解耦评分头层，分别将LLM隐层解码到文本or结构的整个实体空间，并自适应的融合实体概率，实现文本和结构信息的互补预测。通过在多个基准数据集上验证，我们的模型实现了sota的性能，并获得了显著提升。
%
% 创新点：
% 1、一种新的端到端框架，one-shot的全空间预测
% 2、有效平衡结构信息和文本信息：双视角特征表示+解耦预测头
\section{Introduction}
Knowledge Graphs (KGs) serve as pivotal resources for modern artificial intelligence, supporting a multitude of knowledge-intensive tasks \cite{ji2021survey,rossi2021knowledge}, such as question answering \cite{zhai2024towards} and recommendation systems \cite{zhao2024breaking}. However, real-world KGs frequently suffer from incompleteness, necessitating Knowledge Graph Completion (KGC) to predict missing triplets based on observed facts. Recently, Large Language Models (LLMs) have revolutionized Natural Language Processing (NLP), achieving state-of-the-art performance \cite{touvron2023llama,qin2023chatgpt,liu2024deepseek}. Powered by massive parameters that encode vast open-world knowledge, LLMs exhibit exceptional semantic reasoning capabilities. Consequently, leveraging this rich parametric knowledge to enhance KGC has emerged as a promising frontier.\\
Despite this potential, applying LLMs to KGC faces a fundamental granularity mismatch. Tokens are the basic elements for language models, but it needs to take at least several tokens to describe and identify different entities in a KG. Consequently, one category of methods \cite{liu2025filter,wei2023kicgpt,yao2025exploring,chen2024new,jiang2024kg} avoids generation by picking from a limited set of options, sacrificing the LLM's potential for global ranking over the entire entity space and rendering the process inefficient. 
\begin{figure}
    \centering
    \includegraphics[width=1\linewidth]{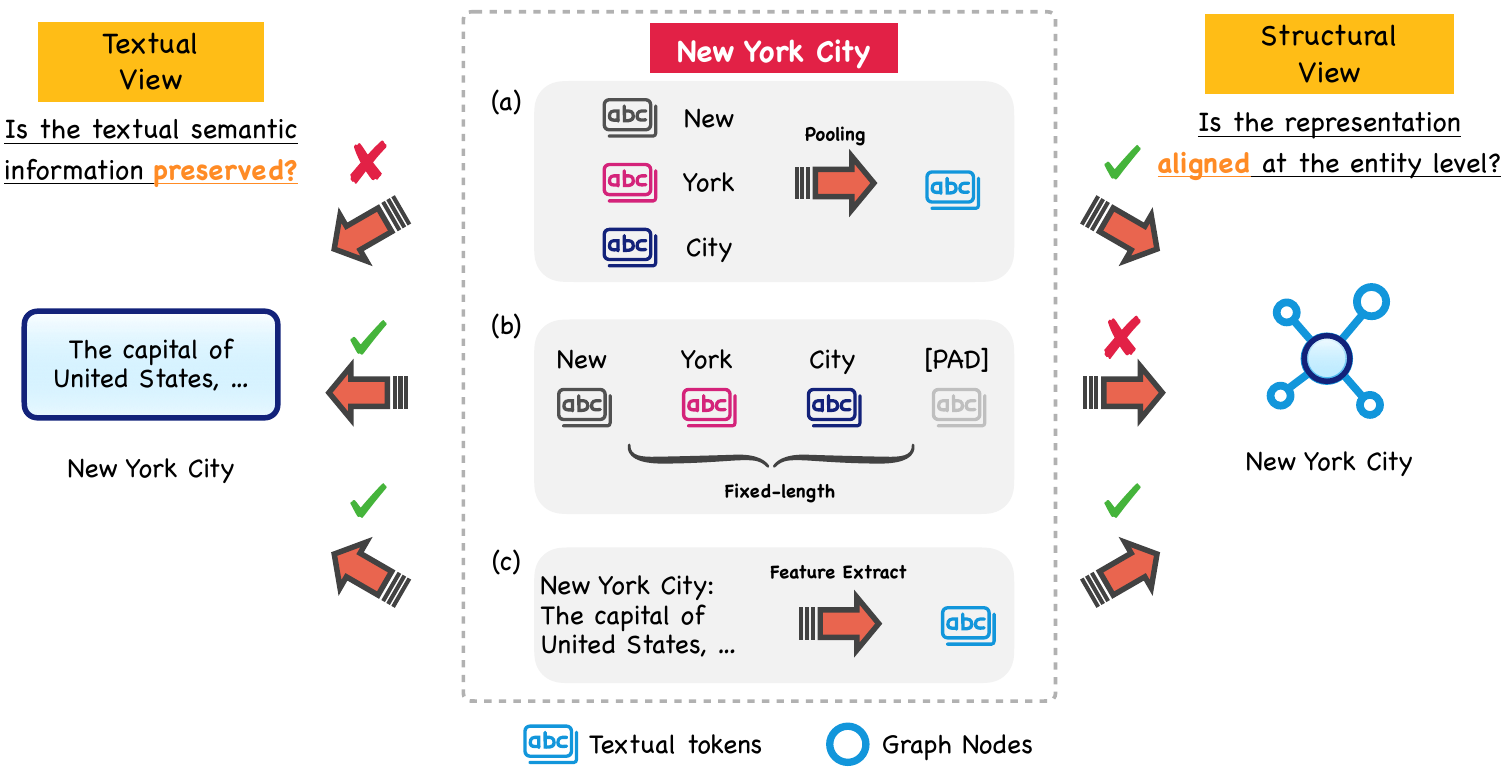}
    \caption{An illustration of existing two strategies of full-space LLM-based methods and KGT. (a) Pooling multiple tokens to unified representations for entities; (b) Decomposing entities into fixed-length sub-word sequences; (c) Constructing feature representations directly at the indivisible entity level.}
    \label{fig:intro}
\end{figure}
To achieve full-space prediction, another category of approaches attempts to align entities with the LLM’s native vocabulary, typically via two strategies: pooling \cite{huang2025elmm,guo2024mkgl} or decomposition \cite{guo2025k} as illustrated in Figure \ref{fig:intro}.\\
However, these two strategies struggle with some significant challenges. Pooling operations compress entity tokens into a single vector, inevitably leading to semantic dilution, which is particularly severe for long or polysemous entities. Besides, decomposing entities into fixed-length sub-word sequences preserves fine-grained semantics but disrupts the structural integrity. Consequently, such methods remain trapped in a trade-off between semantic expressiveness and structural modeling. This compels us to question: \textit{Is following the LLM's native vocabulary truly necessary for full-space prediction on KGs?}\\
To this end, we propose KGT, a novel framework that bridges the granularity mismatch between LLMs and KGs by orchestrating\textbf{ Tokenization, Fusion, and Decoupling}. Specifically, as illustrated in Figure \ref{fig:intro}, we first register entities and relations as special tokens, treating them as indivisible units to maintain granularity consistency. Next, to circumvent the high cost of learning these tokens from scratch, we employ dual-stream specialized token embedding at the input level, where we project pre-trained textual and structural features into a unified latent space and fuse them via a relation-guided gating mechanism to explicitly inject structural priors. Finally, we implement decoupled prediction via dual-view heads, which project the LLM's hidden states into distinct textual and structural subspaces. This design explicitly separates semantic and structural reasoning, producing independent scores that are then adaptively combined via learnable scalers for comprehensive full-space prediction.
Our major contributions are summarized as follows:
\begin{itemize}
    \item  We propose KGT, a novel end-to-end framework that bridges the granularity mismatch via specialized tokenization. By representing entities as indivisible specialized tokens, KGT eliminates the need for candidate filtering, enabling full-space prediction in a single step.
    \item We develop a dual-stream fusion and decoupling architecture. Specifically, we employ a relation-guided gating mechanism to fuse textual and structural features, and dual-view heads to disentangle reasoning. Additionally, utilizing pre-trained embeddings as a warm start ensures superior training efficiency. 
    \item To evaluate the performance of KGT, we conduct comprehensive experiments and further explorations on three public benchmarks. Empirical results demonstrate that KGT outperforms 19 recent baselines, achieving new state-of-the-art results, promoting the stronger capability of LLMs for KGC tasks.
\end{itemize}

\section{Preliminary}
\begin{figure*}[t!]
    \centering
    \includegraphics[width=1\textwidth]{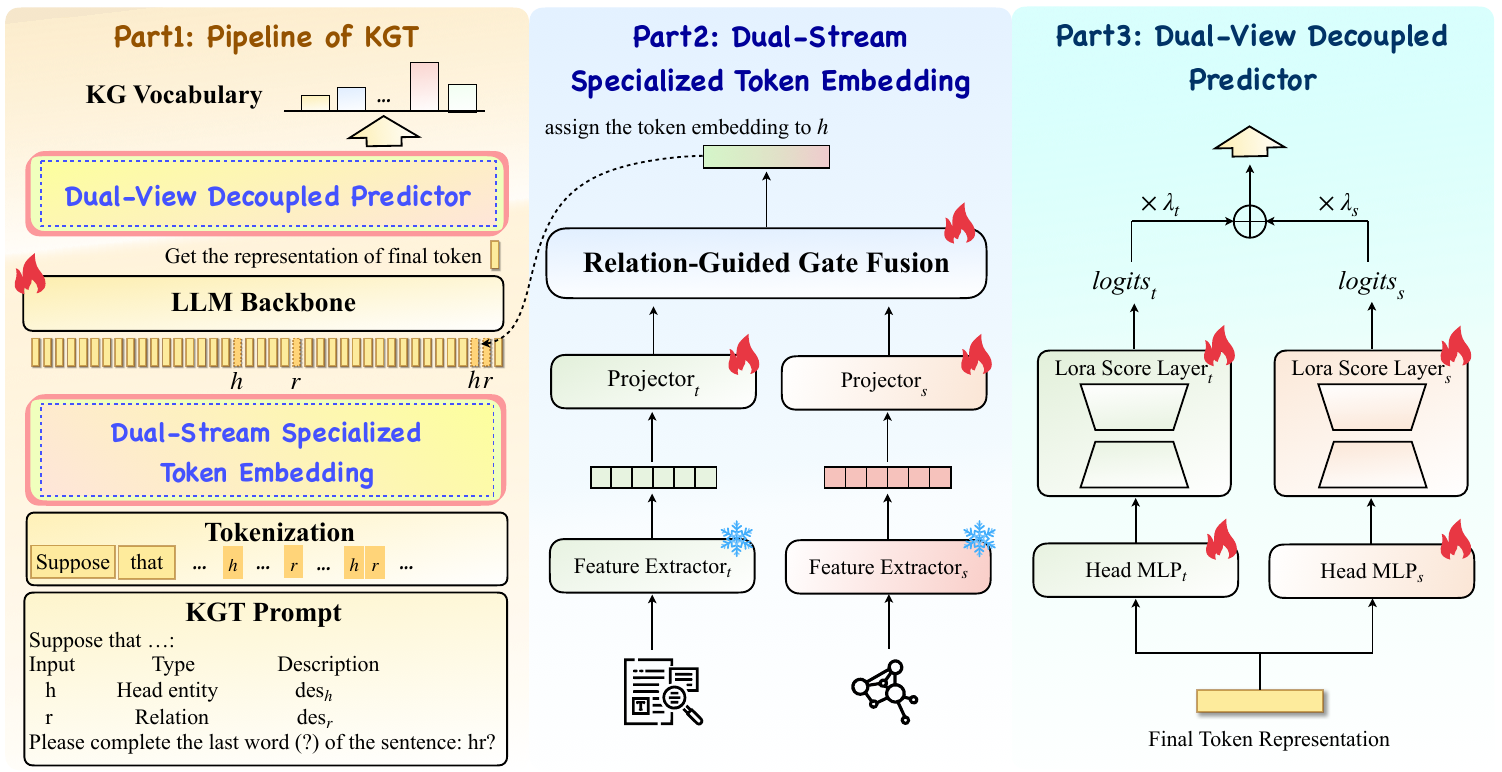}
    \caption{Overview of the KGT framework. \textbf{Part 1} illustrates the overall pipeline of KGT. The tokenizer first processes the input text containing the incomplete triple query, where entities and relations are represented as special tokens added to the original vocabulary. These special tokens obtain their embeddings via the Dual-Stream Specialized Token Embedding module. Subsequently, the LLM Backbone encodes the sequence, extracting the feature of the last token, which is then fed into the Dual-View Decoupled Predictor to generate the probability distribution over the entire entity vocabulary. \textbf{Part 2} details the implementation of the Dual-Stream Specialized Token Embedding, where the dashed line indicates the assignment of the fused specialized feature to the special token representing the head entity h. \textbf{Part 3} depicts the detailed architecture of the Dual-View Decoupled Predictor.} 
    \label{fig:framework}
\end{figure*}
% \subsection{Task Definition}
% A Knowledge Graph is formally defined as $\mathcal{G} = (\mathcal{E}, \mathcal{R}, \mathcal{T})$, where $\mathcal{E}, \mathcal{R}, \mathcal{T}$ denote the entity, relation, and triplet sets, respectively. To encapsulate the textual and structural information of KG elements, we define a set denoted as \(\mathcal{M}=\{t,s\}\), where t and s represent the textual and structural modalities, respectively. For any entity \(e\) and relation \(r\), their raw information corresponding to a modality \(m \in \mathcal{M}\) is denoted as \(\mathcal{X}_m(e)\) and \(\mathcal{X}_m(r)\), respectively. KGC aims to predict the missing tail entity \(t\) given a query $(h, r, ?)$ \cite{bordes2013translating}. We utilize inverse relations \(r^{-1}\) to unify head entity prediction $(?, r, t)$ into the tail prediction format $(t, r^{-1}, ?)$.
% % Each entity and relation possesses textual information and structural information \(\mathbf{X}_m \in \mathbb{R}^{d_t}\) and structural features \(\mathbf{e}_s \in \mathbb{R}^{d_s}\). 
\subsection{Task Definition}
A Knowledge Graph is formally defined as $\mathcal{G} = (\mathcal{E}, \mathcal{R}, \mathcal{T})$, where $\mathcal{E}$, $\mathcal{R}$, and $\mathcal{T}$ denote the sets of entities, relations, and triplets, respectively.  To explicitly model the dual nature of KGs, we define two modalities: textual ($t$) and structural ($s$). For any entity $e \in \mathcal{E}$ (or relation $r \in \mathcal{R}$), we denote its available raw information as $\mathcal{X}_t(e)$ and  $\mathcal{X}_s(e)$, corresponding to the textual semantics and structural topology, respectively.
KGC aims to predict the missing tail entity \(t\) given a query $(h, r, ?)$ \cite{bordes2013translating}. We utilize inverse relations \(r^{-1}\) to unify head entity prediction $(?, r, t)$ into the tail prediction format $(t, r^{-1}, ?)$.
\subsection{Large Language Models}
\label{sec:llm-workflow}
Generally, a typical LLM comprises four key components:\\
\textbf{Tokenizer:} Splits input text into a token sequence \(t_{0:n} \) based on vocabulary \(\mathcal{V}\). In this work, we adopt an expanded tokenizer, registering individual entities and relations as indivisible tokens. For example, entity "Mainz" and relation "capital of" are encoded as tokens <kgl: Mainz> and <kgl: capital of>, respectively; \\
\textbf{Token Embedding: }Maps the discrete tokens to a sequence of low-dimensional vectors \(\mathbf{t}_{0:n}\); \\
\textbf{Transformer:} The core of the LLM, which processes the input embeddings into deep hidden states: 
\begin{equation}
\label{eq:transformer}
\mathbf{h}_{0:n}=\mathcal{M}(\mathbf{t}_{0:n});
\end{equation}
\textbf{Head Layer:} Maps the final hidden state \(\mathbf{h}_n\) to a probability distribution \(\mathbf{p}_{n+1} \in \mathbb{R}^{\lvert \mathcal{V}\rvert}\)  for predicting the next token \(t_{n+1}\):
\begin{equation}
\mathbf{p}_{n+1}=\mathcal{H}(\mathbf{h}_n).
\end{equation}
In this paper, we focus on customizing the \textbf{Token Embedding} and \textbf{Head Layer}. For KG elements added to the vocabulary, we design token embeddings that integrate both textual semantics and structural priors. Distinctively, unlike standard LLMs that generate over the general vocabulary, we re-implement the head layer to map hidden states specifically to the full entity set \(\mathcal{E}\), where \(\mathcal{E}\subset \mathcal{V}\).
\section{Methodology}
In this section, we elaborate on the proposed framework, KGT, in two parts: Dual-Stream Special Token Embedding, Dual-View Decoupled Predictor. Figure \label{fig:framework2} illustrates the overview of KGT. Notably, the prompt in this paper follows that of MKGL \cite{guo2024mkgl}.
\subsection{Dual-Stream Specialized Token Embedding}
Prior research \cite{zhang2024multiple,zhang2025tokenization,cao2022otkge} confirms that textual semantics and structural topology are both indispensable for KGC. However, existing LLM-based methods relying on sub-token composition often compromise or neglect one modality. To effectively model entities and relations, we replace the standard embedding layer with a Dual-Stream Specialized Token Embedding module (\(E_{\text{specialized}}\)). This module treats entities and relations as holistic tokens and processes them via parallel streams of feature extraction,  projection and gating fusion. For clarity, we delineate the process for entities below, noting that relations follow a symmetric procedure.
\paragraph{Feature Extraction.}
\label{para:extraction}
For an entity \(e \in \mathcal{E}\),  we transform its raw inputs $\mathcal{X}_t(e)$ and $\mathcal{X}_s(e)$ into dense feature vectors. For textual information, we employ a sentence embedding extractor (e.g., \textit{text-embedding-3-small} \footnote{https://platform.openai.com/docs/guides/embeddings}) to encode the raw text $\mathcal{X}_t(e)$ into a semantic vector $\mathbf{e}_t \in \mathbb{R}^{d_t}$. Conversely, for the raw structural information, we initialize the structural vector $\mathbf{e}_s \in \mathbb{R}^{d_s}$ using embeddings derived from the TuckER model. 
% The same extraction procedure is applied to relations to obtain $\mathbf{e}_t(r)$ and $\mathbf{e}_s(r)$.
\paragraph{Feature Projection.}
Since the extracted feature vectors \(\mathbf{e}_t\) and \(\mathbf{e}_s\) originate from latent spaces disjoint from the LLM, we employ modality-specific projectors to map them into the LLM's unified semantic space \(\mathbb{R}^d\). Each projector consists of four components: a dropout layer, a fully connected layer, an activation function, and a normalization layer. Specifically, the aligned token embedding \(\mathbf{e}'_t\) and \(\mathbf{e}'_s\) is computed as:
% Specifically, for modality \(m \in \{t, s\}\), the aligned token embedding \(\mathbf{e}'_m\) is computed as:
% \[
% \mathbf{e}'_m = \mathrm{RMSNorm}\left(\sigma\left(\mathbf{W}_m \cdot \mathrm{Dropout}(\mathbf{e}_m)\right)\right)
% \]
\begin{equation}
\label{eq:proj-t}
\mathbf{e}'_t = \mathrm{RMSNorm}\left(\sigma\left(\mathbf{W}_t \cdot \mathrm{Dropout}(\mathbf{e}_t)\right)\right)
\end{equation}
\begin{equation}
\label{eq:proj-s}
\mathbf{e}'_s = \mathrm{RMSNorm}\left(\sigma\left(\mathbf{W}_s \cdot \mathrm{Dropout}(\mathbf{e}_s)\right)\right)
\end{equation}
where \(\mathbf{W}_t \in \mathbb{R}^{d \times d_t}\) and \(\mathbf{W}_s \in \mathbb{R}^{d \times d_s}\) are learnable projection matrix. Note that the bias term is omitted. Consistent with Llama-2, we utilize SiLU \cite{elfwing2018sigmoid} as the activation function \(\sigma\) and LlamaRMSNorm \cite{zhang2019root} for normalization to ensure seamless integration with subsequent layers.  
% It is noteworthy that modality-specific projectors are shared across entities and relations, yielding the aligned relation token embeddings denoted as \(\mathbf{r}'_t\) and \(\mathbf{r}'_s\). 
\paragraph{Relation-Guided Gating Fusion.}
\label{para:fusion}
To effectively capture the varying reliance on textual semantics versus structural topology, we employ a Relation-Guided Gating Fusion (ReGF) mechanism:
\begin{align}
    \label{eq:gate}
    (g_t, g_s) &= \mathrm{Softmax}(z_t, z_s) \\
    \label{eq:noise}
    z_m &= \frac{\mathcal{U}_m(\mathbf{e}'_m) /\sqrt{d} + \delta_m}{\sigma(\epsilon_r)}
\end{align}
Here, \(z_m\) represents the relation-aware logit for modality \(m \in \{t,s\}\), reflecting the relative importance of the textual and structural modalities, respectively.  \(\mathcal{U}_m\) , \(\mathcal{U}'_m\) are two projection layers and the term \(\delta_m \sim \mathcal{N}(0, \mathcal{U}'_m(\mathbf{e}'_m)/\sqrt{d}))\) denotes the tunable Gaussian noise introduced to enhance robustness, which has been proven to work \cite{shazeer2017outrageously}. Furthermore, we introduce a learnable relation-aware temperature \(\epsilon_r\) with a sigmoid function \(\sigma\) to limit the temperature in the range \((0, 1)\). This formulation dynamically calibrates the gating weights according to the relational context before the final fusion
Finally, the entity token embedding \(\mathbf{t}_e\) is derived via soft weighted summation:
\begin{equation}
\label{eq:fusion}
    \mathbf{t}_e = g_t \mathbf{e}'_t + g_s \mathbf{e}'_s
\end{equation}
This fused embedding \(\mathbf{t}_e\) serves as the vector representation for entity tokens in the input sequence \(\mathbf{t}_{0:n}\), as defined in the Preliminary. 
Similarly, the relation token embedding \(\mathbf{t}_r\) is obtained via a symmetric process of feature extraction and gating fusion.
\subsection{Dual-View Decoupled Predictor}
To overcome the limitations of restricted candidate sets inherent in some prior approaches \cite{wei2023kicgpt,liu2025filter}, we propose the Dual-View Decoupled Predictor (\(P_{\text{decoupled}}\)) to achieve one-shot global ranking over the entire entity vocabulary. Additionally, by projecting LLM representations into decoupled textual and structural latent spaces, we effectively utilize the distinct advantages of each modality during the prediction phase.
\paragraph{Dual-View  Head MLPs.}
We employ dual-view head MLPs to process the LLM's final hidden state \(\mathbf{h}_n \in \mathbb{R}^d\) into distinct textual and structural latent representations, denoted as \(\mathbf{h}'_t\in \mathbb{R}^{d_t} \)\  and \(\mathbf{h}'_s\in \mathbb{R}^{d_s}\). Each MLP consists of four components: a dropout layer, a fully connected layer, an activation function, and a normalization layer:
\begin{equation}
\label{eq:head-mlp-t}
\mathbf{h}'_t = \mathrm{RMSNorm}\left(\sigma\left( \mathrm{Dropout}(\mathbf{h}_n)\cdot \mathbf{W}'_{t}\right)\right)
\end{equation}
\begin{equation}
\label{eq:head-mlp-s}
\mathbf{h}'_s = \mathrm{RMSNorm}\left(\sigma\left( \mathrm{Dropout}(\mathbf{h}_n)\cdot \mathbf{W}'_{s}\right)\right)
\end{equation}
where \(\mathbf{W}'_t \in \mathbb{R}^{d \times d_t}\) and \(\mathbf{W}'_s \in \mathbb{R}^{d \times d_s}\) are learnable projection matrices. Consistent with the input projector, we employ SiLU \cite{elfwing2018sigmoid} as the activation function \(\sigma\) and LlamaRMSNorm \cite{zhang2019root} for normalization. It is noteworthy that the bias vector is not used.
% \paragraph{LoRA-based Score Layer.}
% We propose to use low-rank adaptation (LoRA)~\cite{hu2022lora} layers to estimate the probability distribution of all entities. This can be expressed as:
% \begin{align}
% \mathbf{W}^{S}_{m} &= \mathbf{W}_{base, m} + \mathbf{A}_{m}\mathbf{B}_{m} \label{eq:lora_weight} \\
% \mathbf{p}_{m} &= \mathbf{h}'_{m} \mathbf{W}^{S}_{m} \label{eq:score_calc}
% \end{align}
% where \(\mathbf{W}_{base,m}\in \mathbb{R}^{d_m \times |\mathcal{E}|}\) is the frozen scoring matrix initialized by the original entity embeddings, and \(|\mathcal{E}|\) denotes the number of entities in the KG. \(\mathbf{A}_{m} \in \mathbb{R}^{d_m \times r}\) and \(\mathbf{B}_{m} \in \mathbb{R}^{r \times |\mathcal{E}|}\) are the down-projection and up-projection matrices, respectively, with rank \(r \ll d_m\). Following standard practice~\cite{hu2022lora}, \(\mathbf{A}_{m}\) is initialized with random Gaussian noise, while \(\mathbf{B}_{m}\) is initialized to zero. This ensures that the training starts with the original projection logic.
\paragraph{LoRA Score Layer.}
To efficiently adapt the model to the KGC task without full-parameter fine-tuning, we propose a LoRA \cite{hu2022lora} scoring mechanism that leverages the pre-trained entity embeddings as a warm start. We instantiate two separate scoring matrices, \(\mathbf{W}^{S}_t\) and \(\mathbf{W}^{S}_s\), corresponding to the textual and structural streams. Taking the textual modality as an example, the scoring weight is computed as:
\begin{align}
\mathbf{W}^{S}_t &= \mathbf{W}_{base, t} + \mathbf{A}_{t}\mathbf{B}_{t} \label{eq:lora_weight} \\
\mathbf{p}_{t} &= \mathbf{h}'_{t} (\mathbf{W}^{S}_{t})^\top \label{eq:lora-score}
\end{align}
where \(\mathbf{p}_{t} \in \mathbb{R}^{|\mathcal{E}|}\) denotes the prediction logits over the entity vocabulary. Crucially, the frozen base matrix \(\mathbf{W}_{base,t}\in \mathbb{R}^{|\mathcal{E}| \times d_t}\) is initialized with the pre-trained textual entity embeddings defined in the \ref{para:extraction} module, rather than random initialization. Similarly, the structural base matrix \(\mathbf{W}_{base,s}\) is initialized with the TuckER embeddings. \(\mathbf{A}_{t}\) and \(\mathbf{B}_{t}\) are low-rank learnable adapters (\(r \ll d_t\)) initialized with Gaussian noise and zeros, respectively. This design ensures the predictor retains the rich semantic and structural priors captured during the pre-training phase while adapting to the specific ranking objective.
\paragraph{Optimization.}
To effectively combine the predictions from textual and structural views, we employ a learnable logit scaling mechanism (LLS) with two learnable scalar parameters, \(\lambda_t\) and \(\lambda_s\), to dynamically adjust the contribution of each view. The final hybrid logits \(\mathbf{p}_{n+1}\) are computed as the weighted average of the dual-view logits, aligning with the next-token prediction objective:
\begin{equation}
\label{eq:logits-scale}
\mathbf{p}_{n+1} = \frac{1}{2} \left( \lambda_t \mathbf{p}_t + \lambda_s \mathbf{p}_s \right)
\end{equation}
The model is optimized using the standard Cross-Entropy Loss to maximize the likelihood of the ground truth entity \(e^+\):
\begin{equation}
\label{eq:loss}
\mathcal{L} = - \mathbf{p}_{n+1}^{e^+} + \log \left( \sum_{e_j \in \mathcal{E}} \exp(\mathbf{p}_{n+1}^{e_j}) \right)
\end{equation}
where \(\mathbf{p}_{n+1}^e\) denotes the final prediction logit of entity \(e\) as the next token \(t_{n+1}\).
\section{Experiment}

\begin{table*}[!t]
    \centering
    \begin{tabular}{ c  |cc|cc|cccc}
    \toprule

\multirow{2}{*}{\textbf{Methods}} & \multicolumn{2}{c}{\textbf{MKG-W}}&  \multicolumn{2}{c}{\textbf{MKG-Y}}& \multicolumn{4}{c}{\textbf{DB15K}} \\
% \cmidrule(lr){3-4} \cmidrule(lr){5-6} \cmidrule(lr){7-10}
         % \midrule
         &  \textbf{MRR$\uparrow$}&  \textbf{H@1$\uparrow$}&\textbf{MRR$\uparrow$}&  \textbf{H@1$\uparrow$}& \textbf{MRR$\uparrow$}& \textbf{H@1$\uparrow$}& \textbf{H@3$\uparrow$}& \textbf{H@10$\uparrow$}\\
 \midrule
         \textbf{TransE} &  29.19 &  21.06  &30.73 &  23.45 &  24.86 &  12.78 &  31.48 & 47.07 \\
         \textbf{DistMult} &  20.99 &  15.93  &25.04 &  19.33 & 23.03 &  14.78 &  26.28 & 39.59 \\
         \textbf{RotatE} &  33.67 &  26.80  &34.95 &  29.10 &  29.28 &  17.87 &  36.12 & 49.66 \\
         \textbf{TuckER} &  30.39 &  24.44  &37.05 &  34.59 &  33.86 &  25.33 &  37.91 & 50.38 \\
    \midrule
         \textbf{IKRL}&  32.36 &  26.11  &33.22 &  30.37 &  26.82 &  14.09 &  34.93 & 49.09 \\
         \textbf{KG-Bert}& 28.68 & 21.12  &- &  - & 23.94 & 11.98 &31.05  & 46.54\\
         \textbf{FLT-LM}&32.75  &25.89 &- & - & 33.45 &24.56& 37.67 & 50.12\\
 \textbf{OTKGE}& 34.36 & 28.85  &35.51 & 31.97 & 23.86 & 18.45 & 25.89 &34.23 \\
 \textbf{MANS}& 30.88& 24.89 &29.03& 25.25& 28.82& 16.87& 36.58&49.26\\
 \textbf{MMRNS}& 35.03& 28.59 &35.93& 30.53& 32.68& 23.01& 37.86&51.01\\
 \textbf{IMF}& 34.50& 28.77 &35.79& 32.95& 32.25& 24.20& 36.00&48.19\\
 \textbf{VISTA}& 32.91 & 26.12  &30.45 & 24.87 & 30.42 & 22.49 & 33.56 &45.94 \\
 \textbf{AdaMF}& 34.27& 27.21 &38.06& 33.49& 32.51& 21.31& 39.60&51.68\\
 \textbf{MyGO}& 36.10 & 29.78  &\uline{38.44} & 35.01 & 37.72 & 30.08 & 41.26 &52.21 \\
 \textbf{MOMOK}& 35.89& \uline{30.38} &37.91& \uline{35.09} & \uline{39.54} & \uline{32.38} & \uline{43.45} & \uline{54.14} \\
 \midrule
\textbf{ KG-Llama-7b}& -& 20.20 &-& -& -& 13.46& -&-\\
 \textbf{GPT 3.5 Turbo}& -& 22.66 &-& -& -& 21.71& -&-\\
 \textbf{MKGL} $^\clubsuit$& 32.86& 26.54 &29.11& 24.30& 27.14& 18.68& 30.39&43.87\\
 \textbf{K-ON} & \uline{36.64}& 30.05&-& -& 38.10& 30.13& 42.77&53.59\\
 \midrule
 \textbf{KGT}& \textbf{43.27
}& \textbf{36.02} &\textbf{43.62}& \textbf{37.68}& \textbf{42.16}& \textbf{34.06}& \textbf{46.03}& \textbf{57.69}\\ 
 Improvements & +18.1\% & +18.6\%  &+13.5\% & +7.4\% & +6.6\% & +5.2\% & +5.9\% & +6.6\% \\
 \bottomrule
    \end{tabular}
\caption{The main KGC results. $\clubsuit$ represents the experimental results that we reproduced through source code. The best and second-best results are boldfaced and underlined, respectively.  -: unavailable entry.}
\label{tab:main-results}
\end{table*}

\paragraph{Datasets.}
In this paper, we employ three widely recognized MKGC benchmarks DB15K \cite{liu2019mmkg}, MKG-W, and MKG-Y \cite{xu2022relation} to evaluate the model performance. MKG-W and MKG-Y are subsets derived from Wikidata \cite{vrandevcic2014wikidata}, YAGO \cite{suchanek2007yago}, and DBpedia \cite{lehmann2015dbpedia}, respectively. These datasets encompass not only structural triplets but also rich unstructured modalities including text and images. Since LLM-based methods utilize additional textual information, comparing them directly against conventional baselines \cite{yao2025exploring, wei2023kicgpt} leads to unfair evaluations. Therefore, we conduct experiments on multi-modal datasets to guarantee a more rigorous and fairer comparison.  The raw data are obtained from their official release sources. The detailed information on the datasets can be found in Table \ref{tab:dataset} in Appendix \ref{sec:dataset details}. 
\paragraph{Evaluation Protocol.}
Following established protocols \cite{sun2019rotate}, we utilize rank-based metrics including Mean Reciprocal Rank (MRR) and Hits@K ($K=1, 3, 10$) , short for H@K, to evaluate the performance. All results are reported under the filtered setting \cite{bordes2013translating}, which excludes candidate triples existing in the training data for fair comparisons.
\paragraph{Baselines.}
To make a comprehensive performance evaluation, we employ 19 different state-of-the-art MMKGC methods as baselines: the conventional structure-only methods, such as TransE \cite{bordes2013translating}, DistMult \cite{yang2014embedding}, RotatE \cite{sun2019rotate} and Tucker \cite{balavzevic2019tucker}; the methods leveraging external knowledge of image and text modalities, such as IKRL \cite{xie2016image} , TransAE \cite{wang2019multimodal}, KG-Bert \cite{yao2019kg} and FLT-LM \cite{lin2023fusing}, OTKGE \cite{cao2022otkge}, MMRNS \cite{xu2022relation}, VISTA \cite{lee2023vista}, IMF \cite{li2023imf}, AdaMF \cite{zhang2024unleashing}, MyGO \cite{zhang2025tokenization}, MOMOK \cite{zhang2024multiple}; and the LLM-based methods such as KG-Llama7b \cite{yao2025exploring} and GPT 3.5 \cite{zhu2024llms}, MKGL \cite{guo2024mkgl}, K-ON \cite{guo2025k}.
\paragraph{Implementation Details.}
We employ Llama-2-7b-chat \cite{touvron2023llama} as the base LLM model. For parameter-efficient fine-tuning, we set the LoRA rank $r=8$, alpha $\alpha=16$ and dropout$=0.05$. The model is trained on 8 NVIDIA H200 GPUs. Full hyper-parameter configurations are available in Appendix \ref{sec:detail}.
\subsection{Main Results}
% 1、我们可以发现，考虑额外信息的方法总体上比传统的只考虑结构的方法表现更好，这验证了利用外部信息源的有效性。
% 2、部分方法通过结合图像和文本模态，AdaMF、MyGO和MOMOK等模型优于单模态模型,凸显了多模态知识的优势，但是仍存在不少多模态方法甚至表现不如单模态，比如在rotate在mkgw, tucker在mkgy和db15k数据集上优于许多多模态的方法，这说明额外信息并不总是有用的，其存在的噪声可能反而会导致性能下降。
% 3、我们的方法取得了，，，，， 并且显著提升，The main KGT results are detailed in Table \ref{tab:main-results}. Comparison with the recent 18 baselines reveals that KGT makes significant progress in all the metrics and achieves new state-of-the-art results, with an improvement of about 3\%-15\%. 说明了在大模型的优越的推理能力，以及有效融合文本信息和结构信息的重要性。此外我们的方法在mrr和hits@1上的提升都很大，说明In addition, we found that the improvement ratio of the KGT  model for Hits@1 and MRR metrics is usually large, which means that the model achieves the best results in both overall prediction and accurate prediction.
% 4、可以看见早期的方法基于令牌进行优化，性能甚至显著低于non-llm的方法。近两年的方法虽然尝试采用组合llm词表实现了实体级别的优化，但是仍显著低于我们的方法。相比之下，我们的方法直接从实体级别构建特征，有利于结构信息的引入和文本信息的保全。
The main KGT results are detailed in Table \ref{tab:main-results}. We can easily find that methods combining image and text modalities usually show higher performance than conventional structure-only approaches, highlighting the advantages of external knowledge. However, We also notice that some multi-modal methods perform worse than conventional methods on specific datasets. For instance, FLT-LM  \cite{lin2023fusing} and MANS \cite{zhang2023modality} perform worse than RotatE \cite{sun2019rotate} on the MKG-W dataset. This suggests that additional information is not always beneficial, for it may introduce noise to training.\\
In contrast, KGT makes significant progress in all the metrics and achieves new SOTA results, with an improvement of about 5\%-18\%. This demonstrates the superior reasoning capability of our LLM-based framework and the importance of effectively fusing textual and structural information. In addition, we find that the improvement ratio of the KGT for Hits@1 and MRR metrics is usually large, which means that the model achieves the best results in both overall prediction and accurate prediction.\\
Early LLM-based methods, as mentioned in previous sections, are optimized against tokens, resulting in fragmented semantics and performance often inferior to even non-LLM baselines. While recent approaches \cite{guo2024mkgl,guo2025k} have attempted to achieve entity-level optimization by aligning entity with the LLM’s native vocabulary, their performance remains significantly lower than KGT. By contrast, our method constructs features directly at the indivisible entity level, which not only facilitates the seamless injection of structural topology but also maximally preserves the integrity of textual semantics.
\subsection{Computational Cost}
\begin{figure}[t]
    \centering
        \includegraphics[width=0.7\linewidth]{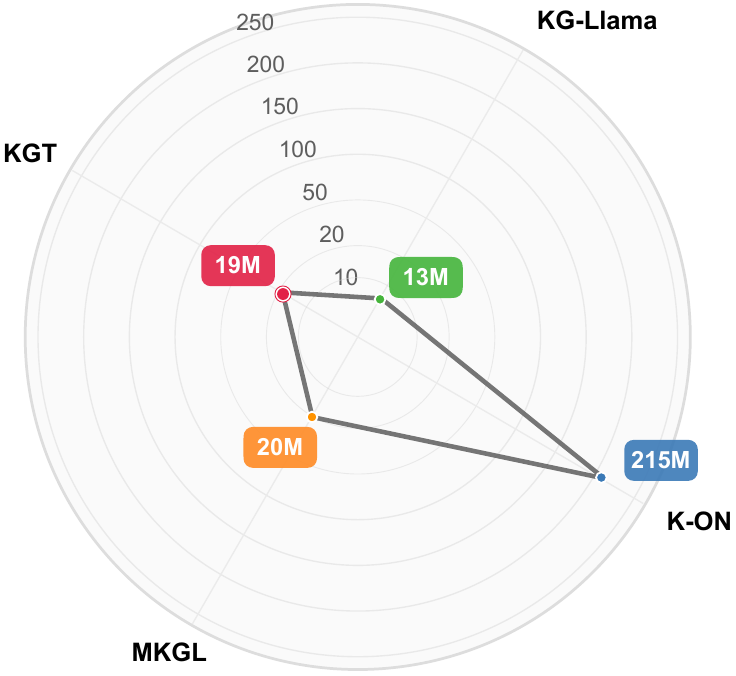}

    \caption{Trainable parameters of some LLM-based KGC methods based on MKG-W.}
    \label{fig:cost}
\end{figure}

\begin{table*}[!htbp]
\centering
\begin{tabular}{c|l|cc|cc|cc}
\toprule 
\multicolumn{2}{c|}{\multirow{2}{*}{\textbf{Setting}}} & \multicolumn{2}{c}{\textbf{MKG-W}} & \multicolumn{2}{c}{\textbf{MKG-Y}} & \multicolumn{2}{c}{\textbf{DB15K}} \\
\multicolumn{2}{c|}{} & \textbf{MRR} & \textbf{H@1} & \textbf{MRR} & \textbf{H@1} & \textbf{MRR} & \textbf{H@1} \\ 
\midrule 
\multicolumn{2}{c|}{Full Model} & 43.27& 36.02& 43.61& 37.68& 42.16& 34.06\\ 
\midrule 
\multirow{2}{*}{\shortstack{Modality\\Contribution}} 
 & (1.1). Structure Modality & 32.73& 27.16& 36.12& 33.39& 37.62& 28.85\\
 & (1.2). Text Modality      & 40.69& 32.44& 40.86& 34.36& 38.08& 30.14\\ 
\midrule 
\multirow{7}{*}{\shortstack{Model\\Design}} 
 & (2.1). w/o structural input     & 41.43& 33.31& 42.15& 35.91& 41.57& 33.47\\
 & (2.2). w/o textual input        & 39.90& 32.78& 39.10& 33.78& 41.74& 33.53\\
 & (2.3). w/o structural predictor& 42.39& 34.81& 41.99& 35.27& 39.53& 32.11\\
 & (2.4). w/o textual predictor   & 33.03& 26.46& 34.20& 31.01& 37.84& 29.06\\
 & (2.5). w/o noise \(\sigma_m\)& 41.05& 33.00& 42.78& 37.23& 41.81& 33.68\\
 & (2.6). w/o relational  \(\epsilon_r\)& 42.63& 35.50& 43.23& 37.29& 41.77& 33.55\\
 & (2.7). w/o LLS& 42.81& 35.33& 42.48& 36.55& 41.93& 33.65\\ 
\bottomrule 
\end{tabular}
\caption{Ablation study of different settings on MKG-W, MKG-Y, and DB15K datasets.}
\label{tab:ablation}
\end{table*}
\begin{figure*}[t]
    \centering
    \includegraphics[width=1\textwidth]{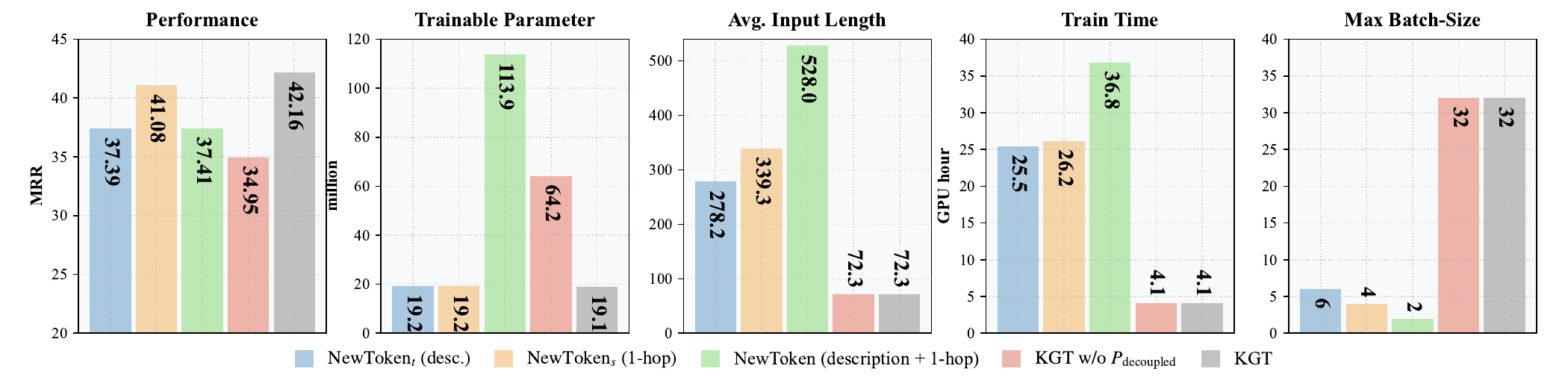}
    \caption{A comprhensive comparison between several viriants of KGT on DB15K.}
    \label{fig:variants}
\end{figure*}
Given that In-Context Learning (ICL) represents an alternative to address the granularity mismatch by supplementing context, we further investigated the rationality and computational efficiency of KGT by comparing it against four variants:
(1) $\text{NewToken}_t(\text{desc.})$, which replaces pre-trained textual features with random initialization while incorporating entity descriptions as context; 
(2) $\text{NewToken}_s(\text{1-hop})$, which replaces pre-trained structural features with random initialization while incorporating 1-hop subgraphs as context; 
(3) $\text{NewToken}(\text{desc. + 1-hop})$, which utilizes randomly initialized tokens with both entity descriptions and 1-hop subgraphs as context. Notably, for this variant, we train both the embedding and the predictor with full parameters to maximize the exploration of ICL effectiveness; 
and (4) KGT w/o $P_{\text{decoupled}}$, which replaces the decoupled predictor $P_{\text{decoupled}}$ with a direct prediction head.\\
As illustrated in Figure \ref{fig:variants}, our method outperforms all variants. First, compared to variants relying on ICL, KGT significantly reduces the average input sequence length and training time by directly leveraging sentence encoders and pre-trained structural features for token initialization. Second, although the $\text{NewToken}_s(\text{1-hop})$ variant lag slightly behind KGT in performance, it suffers from the neighborhood explosion problem. Furthermore, the comparison with KGT w/o $P_{\text{decoupled}}$ validates the rationality and necessity of our dual-head prediction mechanism. This demonstrates that merely injecting knowledge at the input side is insufficient for KGT, necessitating a coordinated output design to fully leverage the initialized representations. Ultimately, KGT supports larger batch sizes during both training and inference phases, enhancing computational efficiency.\\
On the other side, compared to other LLM-based frameworks \cite{guo2024mkgl,guo2025k,touvron2023llama}, our model incurs minimal additional training parameters, even though generating new token embeddings for all entities and relations aligned with the LLM space. The trainable parameters primarily stem from feature alignment, decoupled projections, and the full-space scoring layer. Our approach remains linear in the growth of graph complexity. Figure \ref{fig:cost} illustrates the parameter comparison across various LLM-based KGC methods. When viewed in conjunction with the SOTA KGC performance of KGT in Table \ref{tab:main-results}, we conclude that our method achieves state-of-the-art performance while maintaining high parameter efficiency.
% Compared to other LLM-based frameworks\cite{guo2024mkgl,guo2025k,touvron2023llama}, our model incurs minimal additional training parameters, even though generating new token embeddings for all entities and relations aligned with the LLM space. The trainable parameters primarily stem from feature alignment, decoupled projections, and the full-space scoring layer. Our approach remains linear in the growth of graph complexity. Figure \ref{fig:cost} illustrates the parameter comparison across various LLM-based KGC methods. When viewed in conjunction with the SOTA KGC performance of KGT in Table \ref{tab:main-results}, we conclude that our method achieves state-of-the-art performance while maintaining high parameter efficiency.

% \vspace{-20pt} % 如果底部空白太多，可以用负值减少空白
\subsection{Ablation Studies}
\begin{figure}
    \centering
    \includegraphics[width=\columnwidth]{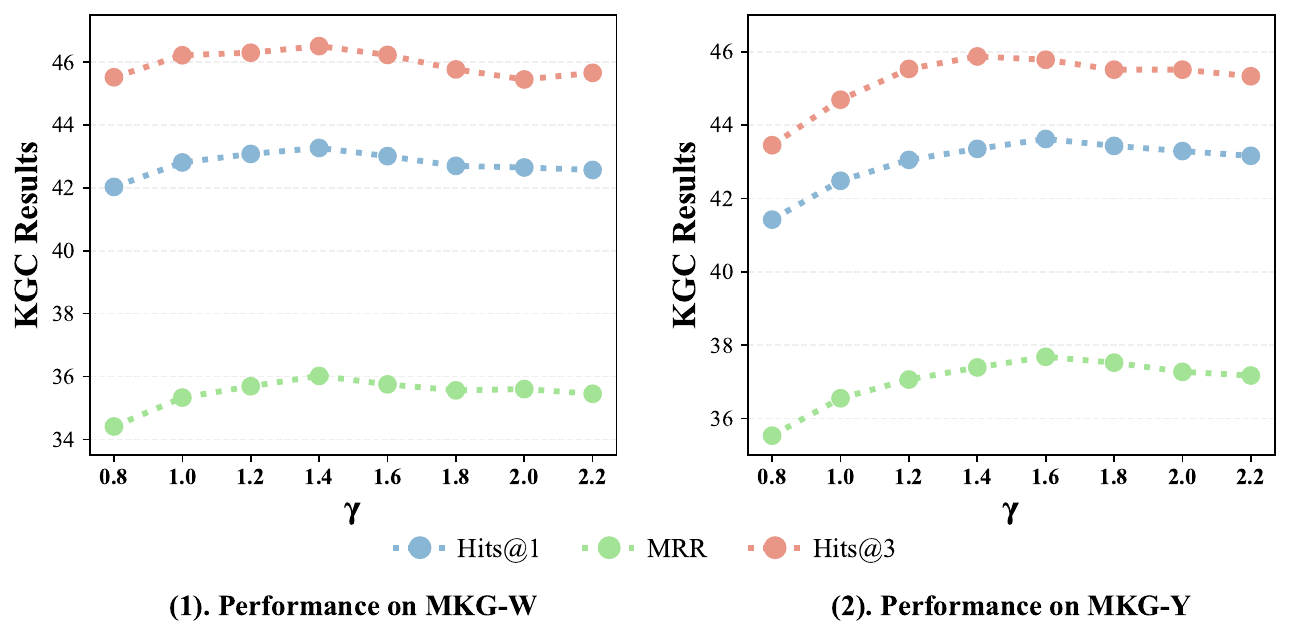}
    \caption{Results of different logits scaling.}
    \label{fig:hyper}
        % \vspace{-10pt} % 如果底部空白太多，可以用负值减少空白
\end{figure}
% 从第一组的实验过中我们可以观察到文本信息和结构信息对最终结果都有贡献，尤其是文本信息，我们认为这与我们使用了高质量的句子特征编码器，结构端只采用了简单的知识表示学习模型，以及大模型对文本信息的优势有关。并且他们都低于全模型的效果。另一方面，我们观察到，即使只使用结构信息，借助大模型强大的推理能力，也能实现优秀的性能。
% 从第二组实验中可以观察到，我们的双流架构，以及输入端带噪声的门控设计和输出端可学习的logits融合都是有效且关键的设计。实验(2.1)-(2.4)证明了双流架构中对称的平衡文本信息和结构信息的有效性，实验（2.5）证明了软门控机制分配模态权重的有效性；（2.6）（2.7）证实了特征融合时关系上下文和可调噪声的有效性，实验（2.7）证明了调节文本logits和结构logits对于最终logits的贡献的有效性。
To confirm the soundness of our design, we conduct further ablation studies to investigate the contribution of modalites and  design in KGT. The experimental results are presented in Table \ref{tab:ablation}.\\
% Our ablation experiments are divided into two main parts. The first part aims to analyze the  structural and textual information and validate whether they positively contribute to the performance. The second part is dedicated to examining our designs in KGT and verifying whether their design has rationality by removing the corresponding modules. 
From the first group of experimental results, we can observe that both textual and structural information positively contribute to the final result. Notably, the textual modality exhibits a dominant influence, which we attribute to the high-quality sentence embeddings and the inherent linguistic advantages of the LLM. \\
% Meanwhile, the structural branch also yields competitive performance despite using a simple embedding model.\\
Moreover, the results from the second group reveal that our key designs significantly contribute to the final performance. Experiments of settings (2.1)–(2.4) confirm the effectiveness of the symmetric dual-stream architecture in balancing textual and structural information. Experiments of settings (2.5) and (2.6) confirm the effectiveness of relational context and tunable noise in the ReGF module. Experiment of setting (2.7) further validates the learnable logit scaling strategy for harmonizing output distributions. \\
% Collectively, these findings indicate that the dual-stream processing combined with the soft gating mechanism has the most profound effect on the final performance, as it effectively arbitrates between semantic and topological signals.\\
We also investigate the effect of the logit scaling coefficient \(\gamma=\lambda_t/\lambda_s\), as depicted in Figure \ref{fig:hyper}. It can be observed that the impact of \(\gamma\) on the final results generally follows a pattern of initial increase followed by a slow decrease. Setting the scaling factor to either too small or too large is detrimental to the model's learning performance. The model achieves the best results with \(\gamma=1.4\) and \(\gamma=1.6\) for MKG-W and MKG-Y, respectively.
\begin{figure}[H]
\small
    \centering
    \includegraphics[width=1\linewidth]{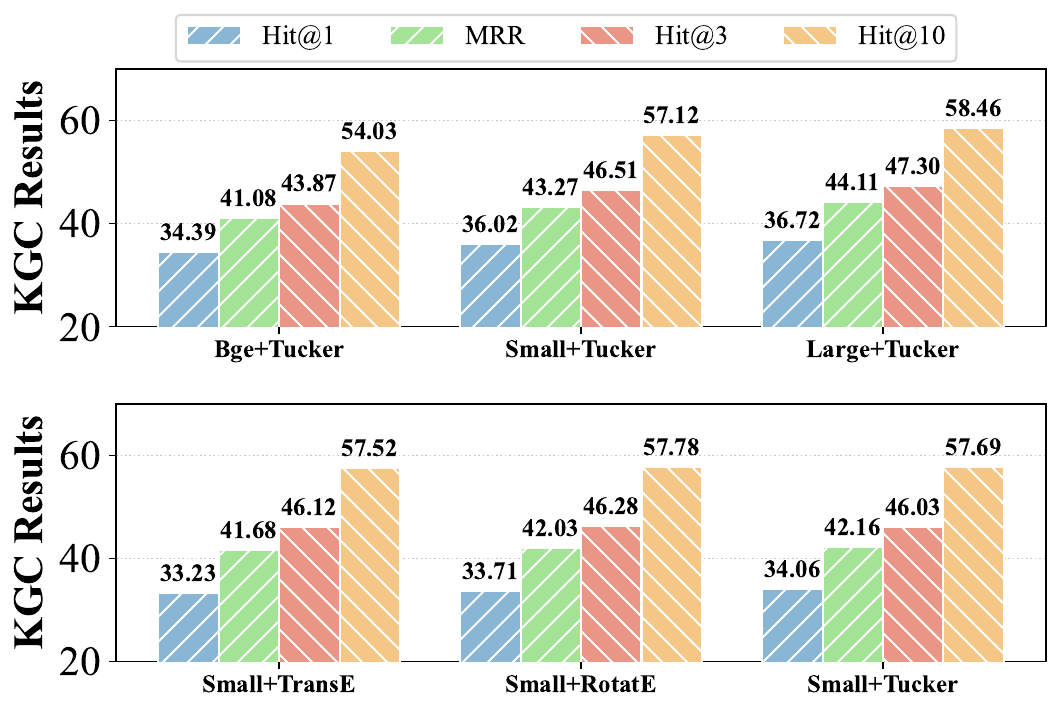}
    \caption{KGC results using different feature extractors. We evaluate diverse textual feature extractors (\textit{text-embedding-3-large}, \textit{bge-large-en-v1.5}) on the MKG-W dataset, and structural feature extractors (TransE, RotatE) on the DB15K dataset.}
    \label{fig:diff-extractor}
\end{figure}
\subsection{Impact of Different Feature Extractors}
% 我们可以看到使用不同的特征提取器对效果有一定的影响，但是即使是最差的结果，也是领先于当前的其他模型的，这也展示了KGT的鲁棒性。另一方面，我们认为这也是符合直觉的，更强的特征提取器代表真更加丰富、有辨识力的特征，经过微调理应获得更优秀的性能。值得注意的是，丰富的特征往往伴随着更大的数据维度，这也带来了更大 的开销，我们将在后面更详细的进行讨论。
To further investigate the impact of different feature extractors on KGT, we conducted experiments using different extraction strategies. The performance results of these combinations are illustrated in Figure \ref{fig:diff-extractor}. We observe that the choice of feature extractors actually does impact performance. However, even the least effective variant surpasses existing SOTA models, underscoring the robustness of KGT. This finding also aligns with our intuition that stronger extractors provide richer and more discriminative features, which naturally lead to superior performance after fine-tuning. Notably, these richer features often come with higher dimensionality and computational overhead.

\subsection{Case Study}
\begin{figure}
    \centering
    \includegraphics[width=1\linewidth]{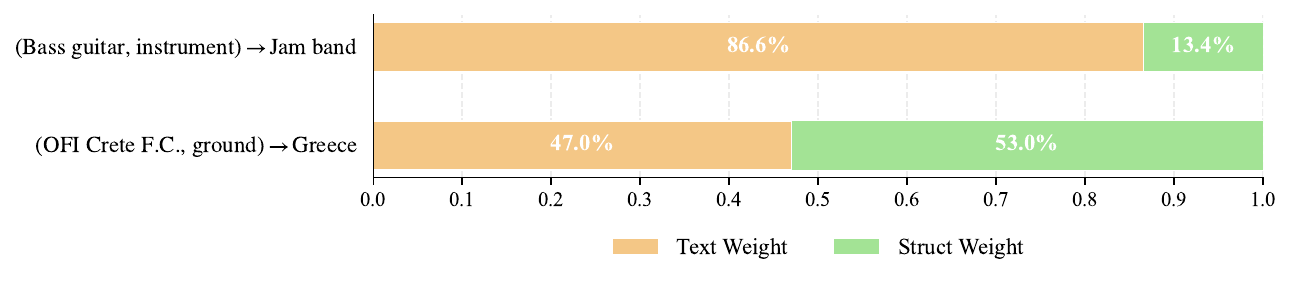}
    \caption{The comparison of model weights among various cases}
    \label{fig:weight}
\end{figure}

To intuitively demonstrate the effectiveness of KGT, we present several representative cases from the DB15K dataset. As shown in Figure  \ref{fig:weight}, KGT adaptively assign modality weights for different triples. Specifically, triples involving geographical topology tend to receive higher structural weights, while those dominated by conceptual semantics are assigned higher textual weights.\\
% For example, triples involving geographical topology tend to receive higher structural weights, while those dominated by conceptual semantics are assigned higher textual weights.
% KGT adaptively assigns modality weights across various triples. Specifically, it allocates higher structural weights to topology-dependent geographical relations, while prioritizing textual weights for semantic-intensive conceptual entities.
% KGT  adaptively assign modality weights for entities and relations, as conceptual entity "Bass Guitar" gets higher textual weight and geographical relation "ground" is assigned higher 
% Specificly, KGT assigns a higher structural weight of 0.53 to the topology-dependent geographical relation "ground", while assigning a higher textual weight of 0.87 to the abstract conceptual entity "Bass guitar".\\
% For the triple (Bass guitar, instrument, Jam band), KGT assigns a dominant weights to the textual modality (0.87) as the relation relies more on abstract semantic concepts. Conversely, for (OFI Crete F.C., ground, Greece), the structural weight rises to 0.53, demonstrating the role of geographic topology.\\
 Table \ref{tab:case_study} presents three representative cases corresponding to structure-dominant, text-dominant, and synergistic scenarios to demonstrate that KGT can rescue a weak modality via the dominant modality's probability distribution. Furthermore, KGT synergizes knowledge from both sides to achieve accurate joint predictions.
% KGT assigns a higher structural weight of 0.53 to the topology-dependent geographical relation "ground", while assigning a higher textual weight of 0.87 to the abstract conceptual entity "Bass guitar".
% \textbf{(a) Text-Dominant Case:} For the triple \textit{(Bass guitar, instrument, Jam band)}, the model assigns a dominant weights to the textual modality (\textbf{0.87}). This is because the relation relies on \textbf{abstract semantic concepts}. The generic entity "Bass guitar" introduces structural noise due to its high connectivity, prompting the model to prioritize the robust parametric knowledge of the LLM.
% \textbf{(b) Structure-Dominant Case:} Conversely, for \textit{(OFI Crete F.C., ground, Greece)}, the structural weight rises to \textbf{0.53}. This demonstrates the model's reliance on \textbf{geographic topology} for long-tail entities. While the LLM may lack specific knowledge about this regional sports club, the KG provides a clear topological cluster of Greek entities, allowing the structural encoder to infer the correct location.
\begin{table}[H]
    \centering
    \caption{Case study experiments on the DB15K dataset}
    \label{tab:case_study}
    
    \fontsize{8pt}{8pt}\selectfont 
    
    \begin{tabularx}{\linewidth}{l|X} 
        \toprule
        \multicolumn{2}{l}{\textbf{Case 1:} ( ? , \textit{composer}, \textit{Adaptation\_(film) })} \\ % 修复：添加了缺失的闭合括号 }
        \midrule
        \textbf{Textual} & \textbf{Rank} of the head entity ``\textit{Carter Burwell}'' : \underline{179} \\
        
        \textbf{Structural} & \textbf{Rank} of the head entity ``\textit{Carter Burwell}'' : \underline{1} \\
        
        \textbf{KGT} & \textbf{Rank} of the head entity ``\textit{Carter Burwell}'' : \underline{1} \\

        \bottomrule \toprule
        \multicolumn{2}{l}{\textbf{Case 2:} ( ? , \textit{distributingCompany}, \textit{Warner\_Music\_Group} )} \\ % 修复：移除了多余的 \textit{) } 及其嵌套错误
        \midrule
        \textbf{Textual} & \textbf{Rank} of the head entity ``\textit{Atlantic\_Records}'' : \underline{1} \\
        
        \textbf{Structural} & \textbf{Rank} of the head entity ``\textit{Atlantic\_Records}'' : \underline{29} \\
        
        \textbf{KGT} & \textbf{Rank} of the head entity ``\textit{Atlantic\_Records}'' : \underline{1} \\
        \midrule % 修复：Case 2 和 3 之间建议加线
        
        \multicolumn{2}{l}{\textbf{Case 3:} ( \textit{New York Stories}, \textit{starring}, ? )} \\ 
        \midrule
        \textbf{Textual} & \textbf{Rank} of the tail entity ``\textit{Mia Farrow}'' :  12 \\
        
        \textbf{Structural} & \textbf{Rank} of the tail entity ``\textit{Mia Farrow}'' : 15 \\
        
        \textbf{KGT} & \textbf{Rank} of the tail entity ``\textit{Mia Farrow}'' : \underline{1} \\
        \bottomrule
    \end{tabularx}
\end{table}

\section{Conclusion}
\label{sec:conculusion}
In this paper, we presented KGT to address the granularity mismatch between LLMs and KGs. Instead of relying on fragmented token sequences, KGT treats entities as indivisible tokens initialized via dual-stream specialized token embedding, allowing for the direct injection of structural priors and warm-start training. Crucially, we introduced a decoupled head predictor to leverage these representations, as our experiments confirm that such specialized inputs should be paired with a corresponding output mechanism. Overall, this end-to-end approach significantly reduces input length and training costs, achieving SOTA performance and promoting the capability of LLMs for KGC tasks.
% \clearpage

\section{Limitations}
Although empirical experiments have confirmed the effectiveness of the proposed KGT, a limitation remains regarding its reliance on pre-computed structural priors. Specifically, since the current framework uses these priors for initialization without joint optimization, it cannot dynamically refine structural features during training. This makes the performance partially dependent on the quality of external KGE models. While employing stronger pre-trained models can mitigate this challenge, it introduces additional computational overhead, which we consider less cost-effective. Consequently, we regard implementing end-to-end structure learning as a more promising avenue. In the future, we will continue to explore this direction, aiming to devise a solution that integrates end-to-end structural learning into our framework. We believe that such an approach will drive further breakthroughs in the performance of LLMs for KG.

\bibliography{custom}

@article{jiang2024kg,
  title={Kg-fit: Knowledge graph fine-tuning upon open-world knowledge},
  author={Jiang, Pengcheng and Cao, Lang and Xiao, Cao Danica and Bhatia, Parminder and Sun, Jimeng and Han, Jiawei},
  journal={Advances in Neural Information Processing Systems},
  volume={37},
  pages={136220--136258},
  year={2024}
}

@inproceedings{chen2024new,
  title={A new pipeline for knowledge graph reasoning enhanced by large language models without fine-tuning},
  author={Chen, Zhongwu and Bai, Long and Li, Zixuan and Huang, Zhen and Jin, Xiaolong and Dou, Yong},
  booktitle={Proceedings of the 2024 Conference on Empirical Methods in Natural Language Processing},
  pages={1366--1381},
  year={2024}
}

@article{huang2025elmm,
  title={ELMM: Efficient Lightweight Multimodal Large Language Models for Multimodal Knowledge Graph Completion},
  author={Huang, Wei and Li, Peining and Liang, Meiyu and Hou, Xu and Du, Junping and Shao, Yingxia and Ye, Guanhua and Liu, Wu and Lu, Kangkang and Yu, Yang},
  journal={arXiv preprint arXiv:2510.16753},
  year={2025}
}

@article{liu2024deepseek,
  title={Deepseek-v3 technical report},
  author={Liu, Aixin and Feng, Bei and Xue, Bing and Wang, Bingxuan and Wu, Bochao and Lu, Chengda and Zhao, Chenggang and Deng, Chengqi and Zhang, Chenyu and Ruan, Chong and others},
  journal={arXiv preprint arXiv:2412.19437},
  year={2024}
}

@article{touvron2023llama,
  title={Llama: Open and efficient foundation language models},
  author={Touvron, Hugo and Lavril, Thibaut and Izacard, Gautier and Martinet, Xavier and Lachaux, Marie-Anne and Lacroix, Timoth{\'e}e and Rozi{\`e}re, Baptiste and Goyal, Naman and Hambro, Eric and Azhar, Faisal and others},
  journal={arXiv preprint arXiv:2302.13971},
  year={2023}
}

@article{qin2023chatgpt,
  title={Is ChatGPT a general-purpose natural language processing task solver?},
  author={Qin, Chengwei and Zhang, Aston and Zhang, Zhuosheng and Chen, Jiaao and Yasunaga, Michihiro and Yang, Diyi},
  journal={arXiv preprint arXiv:2302.06476},
  year={2023}
}

@article{rossi2021knowledge,
  title={Knowledge graph embedding for link prediction: A comparative analysis},
  author={Rossi, Andrea and Barbosa, Denilson and Firmani, Donatella and Matinata, Antonio and Merialdo, Paolo},
  journal={ACM Transactions on Knowledge Discovery from Data (TKDD)},
  volume={15},
  number={2},
  pages={1--49},
  year={2021},
  publisher={ACM New York, NY, USA}
}

@article{ji2021survey,
  title={A survey on knowledge graphs: Representation, acquisition, and applications},
  author={Ji, Shaoxiong and Pan, Shirui and Cambria, Erik and Marttinen, Pekka and Yu, Philip S},
  journal={IEEE transactions on neural networks and learning systems},
  volume={33},
  number={2},
  pages={494--514},
  year={2021},
  publisher={IEEE}
}

@inproceedings{zhao2024breaking,
  title={Breaking the barrier: utilizing large language models for industrial recommendation systems through an inferential knowledge graph},
  author={Zhao, Qian and Qian, Hao and Liu, Ziqi and Zhang, Gong-Duo and Gu, Lihong},
  booktitle={Proceedings of the 33rd ACM International Conference on Information and Knowledge Management},
  pages={5086--5093},
  year={2024}
}

@inproceedings{zhai2024towards,
  title={Towards faithful knowledge graph explanation through deep alignment in commonsense question answering},
  author={Zhai, Weihe and Zubiaga, Arkaitz and Liu, Bingquan and Sun, Cheng-Jie and Zhao, Yalong},
  booktitle={Proceedings of the 2024 Conference on Empirical Methods in Natural Language Processing},
  pages={18920--18930},
  year={2024}
}

@article{guo2024mkgl,
  title={MKGL: mastery of a three-word language},
  author={Guo, Lingbing and Bo, Zhongpu and Chen, Zhuo and Zhang, Yichi and Chen, Jiaoyan and Yarong, Lan and Sun, Mengshu and Zhang, Zhiqiang and Luo, Yangyifei and Li, Qian and others},
  journal={Advances in Neural Information Processing Systems},
  volume={37},
  pages={140509--140534},
  year={2024}
}

@inproceedings{qiu2024joint,
  title={Joint pre-encoding representation and structure embedding for efficient and low-resource knowledge graph completion},
  author={Qiu, Chenyu and Qian, Pengjiang and Wang, Chuang and Yao, Jian and Liu, Li and Wei, Fang and Eddie, Eddie Yk},
  booktitle={Proceedings of the 2024 Conference on Empirical Methods in Natural Language Processing},
  pages={15257--15269},
  year={2024}
}

@article{chen2023dipping,
  title={Dipping plms sauce: Bridging structure and text for effective knowledge graph completion via conditional soft prompting},
  author={Chen, Chen and Wang, Yufei and Sun, Aixin and Li, Bing and Lam, Kwok-Yan},
  journal={arXiv preprint arXiv:2307.01709},
  year={2023}
}

@inproceedings{liu2022know,
  title={I know what you do not know: Knowledge graph embedding via co-distillation learning},
  author={Liu, Yang and Sun, Zequn and Li, Guangyao and Hu, Wei},
  booktitle={Proceedings of the 31st ACM international conference on information \& knowledge management},
  pages={1329--1338},
  year={2022}
}

@inproceedings{wang2021structure,
  title={Structure-augmented text representation learning for efficient knowledge graph completion},
  author={Wang, Bo and Shen, Tao and Long, Guodong and Zhou, Tianyi and Wang, Ying and Chang, Yi},
  booktitle={Proceedings of the web conference 2021},
  pages={1737--1748},
  year={2021}
}

@article{chen2022knowledge,
  title={Knowledge is flat: A seq2seq generative framework for various knowledge graph completion},
  author={Chen, Chen and Wang, Yufei and Li, Bing and Lam, Kwok-Yan},
  journal={arXiv preprint arXiv:2209.07299},
  year={2022}
}

@article{wang2022simkgc,
  title={Simkgc: Simple contrastive knowledge graph completion with pre-trained language models},
  author={Wang, Liang and Zhao, Wei and Wei, Zhuoyu and Liu, Jingming},
  journal={arXiv preprint arXiv:2203.02167},
  year={2022}
}

@inproceedings{daza2021inductive,
  title={Inductive entity representations from text via link prediction},
  author={Daza, Daniel and Cochez, Michael and Groth, Paul},
  booktitle={Proceedings of the Web Conference 2021},
  pages={798--808},
  year={2021}
}

@inproceedings{zhang2020pretrain,
  title={Pretrain-KGE: Learning knowledge representation from pretrained language models},
  author={Zhang, Zhiyuan and Liu, Xiaoqian and Zhang, Yi and Su, Qi and Sun, Xu and He, Bin},
  booktitle={Findings of the association for computational linguistics: EMNLP 2020},
  pages={259--266},
  year={2020}
}

@inproceedings{luo2025gltw,
  title={Gltw: Joint improved graph transformer and llm via three-word language for knowledge graph completion},
  author={Luo, Kangyang and Bai, Yuzhuo and Gao, Cheng and Si, Shuzheng and Liu, Zhu and Shen, Yingli and Wang, Zhitong and Kong, Cunliang and Li, Wenhao and Huang, Yufei and others},
  booktitle={Findings of the Association for Computational Linguistics: ACL 2025},
  pages={11328--11344},
  year={2025}
}

@inproceedings{zhang2024making,
  title={Making large language models perform better in knowledge graph completion},
  author={Zhang, Yichi and Chen, Zhuo and Guo, Lingbing and Xu, Yajing and Zhang, Wen and Chen, Huajun},
  booktitle={Proceedings of the 32nd ACM international conference on multimedia},
  pages={233--242},
  year={2024}
}

@inproceedings{liu2025filter,
  title={Filter-then-generate: Large language models with structure-text adapter for knowledge graph completion},
  author={Liu, Ben and Zhang, Jihai and Lin, Fangquan and Yang, Cheng and Peng, Min},
  booktitle={Proceedings of the 31st International Conference on Computational Linguistics},
  pages={11181--11195},
  year={2025}
}

@inproceedings{wei2023kicgpt,
  title={Kicgpt: Large language model with knowledge in context for knowledge graph completion},
  author={Wei, Yanbin and Huang, Qiushi and Zhang, Yu and Kwok, James},
  booktitle={Findings of the association for computational linguistics: EMNLP 2023},
  pages={8667--8683},
  year={2023}
}

@article{zhang2019root,
  title={Root mean square layer normalization},
  author={Zhang, Biao and Sennrich, Rico},
  journal={Advances in neural information processing systems},
  volume={32},
  year={2019}
}

@article{elfwing2018sigmoid,
  title={Sigmoid-weighted linear units for neural network function approximation in reinforcement learning},
  author={Elfwing, Stefan and Uchibe, Eiji and Doya, Kenji},
  journal={Neural networks},
  volume={107},
  pages={3--11},
  year={2018},
  publisher={Elsevier}
}

@article{shazeer2017outrageously,
  title={Outrageously large neural networks: The sparsely-gated mixture-of-experts layer},
  author={Shazeer, Noam and Mirhoseini, Azalia and Maziarz, Krzysztof and Davis, Andy and Le, Quoc and Hinton, Geoffrey and Dean, Jeff},
  journal={arXiv preprint arXiv:1701.06538},
  year={2017}
}

@inproceedings{guo2025k,
  title={K-ON: Stacking Knowledge On the Head Layer of Large Language Model},
  author={Guo, Lingbing and Zhang, Yichi and Bo, Zhongpu and Chen, Zhuo and Sun, Mengshu and Zhang, Zhiqiang and Zhang, Wen and Chen, Huajun},
  booktitle={Proceedings of the AAAI Conference on Artificial Intelligence},
  volume={39},
  number={11},
  pages={11745--11753},
  year={2025}
}

@article{hu2022lora,
  title={Lora: Low-rank adaptation of large language models.},
  author={Hu, Edward J and Shen, Yelong and Wallis, Phillip and Allen-Zhu, Zeyuan and Li, Yuanzhi and Wang, Shean and Wang, Lu and Chen, Weizhu and others},
  journal={ICLR},
  volume={1},
  number={2},
  pages={3},
  year={2022}
}

@article{zhang2024unleashing,
  title={Unleashing the power of imbalanced modality information for multi-modal knowledge graph completion},
  author={Zhang, Yichi and Chen, Zhuo and Liang, Lei and Chen, Huajun and Zhang, Wen},
  journal={arXiv preprint arXiv:2402.15444},
  year={2024}
}

@inproceedings{zhang2025tokenization,
  title={Tokenization, fusion, and augmentation: Towards fine-grained multi-modal entity representation},
  author={Zhang, Yichi and Chen, Zhuo and Guo, Lingbing and Xu, Yajing and Hu, Binbin and Liu, Ziqi and Zhang, Wen and Chen, Huajun},
  booktitle={Proceedings of the AAAI Conference on Artificial Intelligence},
  volume={39},
  number={12},
  pages={13322--13330},
  year={2025}
}

@article{zhang2024multiple,
  title={Multiple heads are better than one: Mixture of modality knowledge experts for entity representation learning},
  author={Zhang, Yichi and Chen, Zhuo and Guo, Lingbing and Xu, Yajing and Hu, Binbin and Liu, Ziqi and Zhang, Wen and Chen, Huajun},
  journal={arXiv preprint arXiv:2405.16869},
  year={2024}
}

@inproceedings{lee2023vista,
  title={VISTA: Visual-textual knowledge graph representation learning},
  author={Lee, Jaejun and Chung, Chanyoung and Lee, Hochang and Jo, Sungho and Whang, Joyce},
  booktitle={Findings of the association for computational linguistics: EMNLP 2023},
  pages={7314--7328},
  year={2023}
}

@article{cao2022otkge,
  title={Otkge: Multi-modal knowledge graph embeddings via optimal transport},
  author={Cao, Zongsheng and Xu, Qianqian and Yang, Zhiyong and He, Yuan and Cao, Xiaochun and Huang, Qingming},
  journal={Advances in neural information processing systems},
  volume={35},
  pages={39090--39102},
  year={2022}
}

@inproceedings{balavzevic2019tucker,
  title={Tucker: Tensor factorization for knowledge graph completion},
  author={Bala{\v{z}}evi{\'c}, Ivana and Allen, Carl and Hospedales, Timothy},
  booktitle={Proceedings of the 2019 conference on empirical methods in natural language processing and the 9th international joint conference on natural language processing (EMNLP-IJCNLP)},
  pages={5185--5194},
  year={2019}
}

@article{yang2014embedding,
  title={Embedding entities and relations for learning and inference in knowledge bases},
  author={Yang, Bishan and Yih, Wen-tau and He, Xiaodong and Gao, Jianfeng and Deng, Li},
  journal={arXiv preprint arXiv:1412.6575},
  year={2014}
}

@article{zhu2024llms,
  title={Llms for knowledge graph construction and reasoning: Recent capabilities and future opportunities},
  author={Zhu, Yuqi and Wang, Xiaohan and Chen, Jing and Qiao, Shuofei and Ou, Yixin and Yao, Yunzhi and Deng, Shumin and Chen, Huajun and Zhang, Ningyu},
  journal={World Wide Web},
  volume={27},
  number={5},
  pages={58},
  year={2024},
  publisher={Springer}
}

@inproceedings{yao2025exploring,
  title={Exploring large language models for knowledge graph completion},
  author={Yao, Liang and Peng, Jiazhen and Mao, Chengsheng and Luo, Yuan},
  booktitle={ICASSP 2025-2025 IEEE International Conference on Acoustics, Speech and Signal Processing (ICASSP)},
  pages={1--5},
  year={2025},
  organization={IEEE}
}

@article{lin2023fusing,
  title={Fusing topology contexts and logical rules in language models for knowledge graph completion},
  author={Lin, Qika and Mao, Rui and Liu, Jun and Xu, Fangzhi and Cambria, Erik},
  journal={Information Fusion},
  volume={90},
  pages={253--264},
  year={2023},
  publisher={Elsevier}
}

@inproceedings{wang2019multimodal,
  title={Multimodal data enhanced representation learning for knowledge graphs},
  author={Wang, Zikang and Li, Linjing and Li, Qiudan and Zeng, Daniel},
  booktitle={2019 International Joint Conference on Neural Networks (IJCNN)},
  pages={1--8},
  year={2019},
  organization={IEEE}
}

@article{yao2019kg,
  title={KG-BERT: BERT for knowledge graph completion},
  author={Yao, Liang and Mao, Chengsheng and Luo, Yuan},
  journal={arXiv preprint arXiv:1909.03193},
  year={2019}
}

@article{sun2019rotate,
  title={Rotate: Knowledge graph embedding by relational rotation in complex space},
  author={Sun, Zhiqing and Deng, Zhi-Hong and Nie, Jian-Yun and Tang, Jian},
  journal={arXiv preprint arXiv:1902.10197},
  year={2019}
}

@article{xie2016image,
  title={Image-embodied knowledge representation learning},
  author={Xie, Ruobing and Liu, Zhiyuan and Luan, Huanbo and Sun, Maosong},
  journal={arXiv preprint arXiv:1609.07028},
  year={2016}
}

@inproceedings{liu2019mmkg,
  title={MMKG: multi-modal knowledge graphs},
  author={Liu, Ye and Li, Hui and Garcia-Duran, Alberto and Niepert, Mathias and Onoro-Rubio, Daniel and Rosenblum, David S},
  booktitle={European Semantic Web Conference},
  pages={459--474},
  year={2019},
  organization={Springer}
}

@article{vrandevcic2014wikidata,
  title={Wikidata: a free collaborative knowledgebase},
  author={Vrande{\v{c}}i{\'c}, Denny and Kr{\"o}tzsch, Markus},
  journal={Communications of the ACM},
  volume={57},
  number={10},
  pages={78--85},
  year={2014},
  publisher={ACM New York, NY, USA}
}

@inproceedings{suchanek2007yago,
  title={Yago: a core of semantic knowledge},
  author={Suchanek, Fabian M and Kasneci, Gjergji and Weikum, Gerhard},
  booktitle={Proceedings of the 16th international conference on World Wide Web},
  pages={697--706},
  year={2007}
}

@article{lehmann2015dbpedia,
  title={Dbpedia--a large-scale, multilingual knowledge base extracted from wikipedia},
  author={Lehmann, Jens and Isele, Robert and Jakob, Max and Jentzsch, Anja and Kontokostas, Dimitris and Mendes, Pablo N and Hellmann, Sebastian and Morsey, Mohamed and Van Kleef, Patrick and Auer, S{\"o}ren and others},
  journal={Semantic web},
  volume={6},
  number={2},
  pages={167--195},
  year={2015},
  publisher={SAGE Publications Sage UK: London, England}
}

@article{bordes2013translating,
  title={Translating embeddings for modeling multi-relational data},
  author={Bordes, Antoine and Usunier, Nicolas and Garcia-Duran, Alberto and Weston, Jason and Yakhnenko, Oksana},
  journal={Advances in neural information processing systems},
  volume={26},
  year={2013}
}

@inproceedings{zhang2023modality,
  title={Modality-aware negative sampling for multi-modal knowledge graph embedding},
  author={Zhang, Yichi and Chen, Mingyang and Zhang, Wen},
  booktitle={2023 International Joint Conference on Neural Networks (IJCNN)},
  pages={1--8},
  year={2023},
  organization={IEEE}
}

@inproceedings{li2023imf,
  title={IMF: interactive multimodal fusion model for link prediction},
  author={Li, Xinhang and Zhao, Xiangyu and Xu, Jiaxing and Zhang, Yong and Xing, Chunxiao},
  booktitle={Proceedings of the ACM web conference 2023},
  pages={2572--2580},
  year={2023}
}

@inproceedings{xu2022relation,
  title={Relation-enhanced negative sampling for multimodal knowledge graph completion},
  author={Xu, Derong and Xu, Tong and Wu, Shiwei and Zhou, Jingbo and Chen, Enhong},
  booktitle={Proceedings of the 30th ACM international conference on multimedia},
  pages={3857--3866},
  year={2022}
}

@article{kau2024combining,
  title={Combining knowledge graphs and large language models},
  author={Kau, Amanda and He, Xuzeng and Nambissan, Aishwarya and Astudillo, Aland and Yin, Hui and Aryani, Amir},
  journal={arXiv preprint arXiv:2407.06564},
  year={2024}
}

@article{sun2023think,
  title={Think-on-graph: Deep and responsible reasoning of large language model on knowledge graph},
  author={Sun, Jiashuo and Xu, Chengjin and Tang, Lumingyuan and Wang, Saizhuo and Lin, Chen and Gong, Yeyun and Ni, Lionel M and Shum, Heung-Yeung and Guo, Jian},
  journal={arXiv preprint arXiv:2307.07697},
  year={2023}
}

@inproceedings{dettmers2018convolutional,
  title={Convolutional 2d knowledge graph embeddings},
  author={Dettmers, Tim and Minervini, Pasquale and Stenetorp, Pontus and Riedel, Sebastian},
  booktitle={Proceedings of the AAAI conference on artificial intelligence},
  volume={32},
  number={1},
  year={2018}
}

\clearpage
\appendix
\label{sec:appendix}
\section{Related Works}
\paragraph{Knowledge Graph Completion}  
Knowledge Graph (KG) completion is one of the most important tasks in the KG area, with mainstream methods generally falling into two categories: embedding-based and text-based approaches. Embedding-based methods (\cite{bordes2013translating,sun2019rotate,yang2014embedding,balavzevic2019tucker}) generate low-dimensional vectors to model connectivity patterns via scoring functions. While simple and effective, they focus on relational information but often ignore the rich contextual information embedded within the text. Conversely, text-based methods (\cite{yao2019kg,daza2021inductive,chen2022knowledge,wang2022simkgc}) utilize Pre-trained Language Models (PLMs) to encode textual descriptions, often introducing contrastive learning to enhance discriminative ability. However, these methods lack the inherent structural knowledgeof KGs. Consequently, recent efforts\cite{zhang2020pretrain,qiu2024joint,chen2023dipping,liu2022know,wang2021structure} have attempted to combine embedding- and text-based paradigms to leverage their complementary strengths for superior performance.
\paragraph{LLM-based Knowledge Graph Completion.}
Due to rich semantic features, LLMs are deemed highly promising in the realm of KGC. However, the natural gap between the graph structure of KGs and the natural language makes it difficult to apply LLMs directly to KGC tasks. For instance, methods like KGLlama~\cite{yao2025exploring} and KoPA~\cite{zhang2024making} simplify the task into triplet classification, i.e., estimating the correctness of a given triplet. Subsequently, many approaches attempt to complete the task within restricted candidate sets. Some works (\cite{wei2023kicgpt,sun2023think,kau2024combining}) employ prompt engineering or retrieval strategies to obtain KG information and input it into LLMs for entity reranking. However, such an approach is evidently suboptimal. Recently, some methods\cite{guo2025k,guo2024mkgl,luo2025gltw} have explored how to achieve full-space prediction. For example, K-ON\cite{guo2025k} integrates KG knowledge into the LLM by employing multiple head layers to predict multi-step tokens simultaneously, and aggregates the multi-step prediction tokens into an entity probability. Nevertheless, methods for full-space prediction are still scarce, and specifically, the effective synergy of textual and structural information remains an open problem.

\section{Implementation Details}
\label{sec:detail}

We introduce Algorithm \ref{alg:kgt} to demonstrate the fine-tuning process of KGT for KGC. The main hyper-parameter settings are summarized in Table \ref{tab:hyper_parameter}. Besides, we also provide the prompt we used in Table \ref{tab:kg_prompt}.
\begin{algorithm*}[t]
\caption{KGT for KGC}
\label{alg:kgt}
\begin{algorithmic}[1]
\State \textbf{Input:} the training KG $\mathcal{G}$, the language model $\mathcal{M}$, the original vocabulary of the LLM \(\mathcal{V}_{\text{llm}}\), the original token embedding layer of the LLM \(\mathbf{T}\), dual-stream specialized token embedding \(E_{\text{specialized}}\), and decoupled full-space predictor \(P_{\text{decoupled}}\).

\For{\textbf{each} batched triplets in the training KG $\mathcal{G}$}
    \State Construct and tokenize the input instructions; 
    \For{\textbf{each} input token sequence \( t_{0:n}\)}
        \For{\textbf{each} token \(t_{k}\)}
            \If{\(t_{k}\in \mathcal{V}_{\text{llm}}\)}
                \State \(\mathbf{t}_k \gets \mathbf{T}(t_k)\);      
            \Else
                \State \(\mathbf{t}_k \gets E_{\text{specialized}}(t_k)\) (Equations (\ref{eq:proj-t}--\ref{eq:fusion}));
            \EndIf
        \EndFor
    \EndFor
    
    \State \(\mathbf{h}_{0:n}\gets \mathcal{M}(\mathbf{t}_{0:n})\), obtaining the output hidden states of LLM (Equation (\ref{eq:transformer}));
    \State Compute KGT textual and structural predictions with \(P_{\text{decoupled}}\) (Equations (\ref{eq:head-mlp-t}--\ref{eq:lora-score}));
    \State Compute the final full-space prediction \(\mathbf{p}_{n+1}\) following (Equation (\ref{eq:logits-scale}));
    \State Compute and minimize the cross-entropy loss \(\mathcal{L}\) (Equation (\ref{eq:loss}));
\EndFor
\end{algorithmic}
\end{algorithm*}

\begin{table*}[t]
\centering
\small
\setlength{\tabcolsep}{2pt}
\begin{tabularx}{\textwidth}{c *{7}{>{\centering\arraybackslash}X}}
\toprule
\textbf{Dataset} & \textbf{LLM} & \textbf{\makecell{text\\dim}} & \textbf{\makecell{struct\\dim}} & \textbf{\makecell{LoRA\\\(r\)}} & \textbf{\makecell{LoRA\\\(\alpha\)}} & \textbf{\makecell{LORA target\\modules}} & \textbf{\makecell{logits scaling\\ratio \(\gamma\)}} \\
\midrule
\textbf{MKG-W} & \makecell{Llama-2-7b} & 1536 & 256 & 8 & 16 & query, value & 1.4 \\
\textbf{MKG-Y} & \makecell{Llama-2-7b} & 1536 & 256 & 8 & 16 & query, value & 1.6 \\
\textbf{DB15K} & \makecell{Llama-2-7b} & 1536 & 256 & 8 & 16 & query, value & 1.5 \\
\midrule
\midrule
\textbf{Dataset} & \textbf{epoch} & \textbf{\makecell{batch size\\per device}} & \textbf{\makecell{gradient\\accumulation\\steps}} & \textbf{\makecell{learning\\rate}} & \textbf{optimizer} & \textbf{\makecell{text\\dropout}} & \textbf{\makecell{struct\\dropout}} \\
\midrule
\textbf{MKG-W} & 8 & 32 & 1 & 2e-4 & \makecell{Adamw\_8bit} & 0.2 & 0.4 \\
\textbf{MKG-Y} & 8 & 32 & 1 & 2e-4 & \makecell{Adamw\_8bit} & 0.1 & 0.3 \\
\textbf{DB15K} & 8 & 32 & 1 & 4e-4 & \makecell{Adamw\_8bit} & 0.0 & 0.5 \\
\bottomrule
\end{tabularx}
\caption{Hyper-parameter settings in the main experients.}
\label{tab:hyper_parameter}
\end{table*}
\begin{table}[t]
\centering
\captionsetup{justification=centering} 

\begin{tabularx}{\textwidth}{X}
\toprule
Suppose that you are an excellent linguist studying a new three-word language for knowledge graph. Given the following dictionary: \\
\vspace{-0.5em} 
\begin{tabular}{@{}lll} 
Input & Type & Description \\
$<$kgl:\textit{Mainz}$>$& Head entity & Mainz\\
$<$kgl:\textit{capital of}$>$& Relation & capital of\\
\end{tabular} \\
\vspace{-0.5em} 
Please complete the last word (?) of the sentence: $<$kgl:\textit{Mainz}$>$$<$kgl:\textit{capital of}$>$? \\
\bottomrule
\caption{The prompt of KGT. Note that we take entity "Mainz" and relation "capital of" as a example.}
\label{tab:kg_prompt}
\end{tabularx}
\end{table}

 \section{Dataset Datails}
\label{sec:dataset details}
\label{tab:dataset}

\begin{table}[H]
\centering
% 使用 resizebox 强制调整宽度为行宽
\resizebox{\linewidth}{!}{%
    \begin{tabular}{c|ccccc|c}
    \toprule
    {\textbf{Dataset}} & {\textbf{\#Entity}} & {\textbf{\#Relation}} & {\textbf{\#Train}} & {\textbf{\#Valid}} & {\textbf{\#Test}} & \textbf{Text} \\
    \midrule
    \textbf{MKG-W} & 15000 & 169 & 34196 & 4276 & 4274 & 14123 \\
    \textbf{MKG-Y} & 15000 & 28 & 21310 & 2665 & 2663 & 12305 \\
    \textbf{DB15K} & 12842 & 279 & 79222 & 9902 & 9904 & 9078 \\
    \bottomrule
    \end{tabular}%
}
\caption{Statistical information of the three datasets in our experiments. The entity descriptions are provided by the original datasets.}
\label{tab:dataset}
\end{table}

\section{Additional Experiment Results}
\label{sec:add-exp}
We present the additional experimental results in this section. 

\subsection{Full results on MKG-W and MKG-Y} 
In our experiments, we present the MRR and Hits@1 results in Table \ref{tab:main-results}. We now present the results for the complete set of four metrics in both datasets in the Table \ref{tab:full-results}.
\begin{table*}
    \centering
    \begin{tabular}{ c  |cccc|cccc}
    \toprule

\multirow{2}{*}{\textbf{Methods}} & \multicolumn{4}{c|}{\textbf{MKG-W}}& \multicolumn{4}{c}{\textbf{MKG-Y}}\\
% \cmidrule(lr){3-4} \cmidrule(lr){5-6} \cmidrule(lr){7-10}
         % \midrule
         &  \textbf{MRR} &  \textbf{H@1}  & \textbf{H@3}&\textbf{H@10}&  \textbf{MRR} &  \textbf{H@1} & \textbf{H@3}&\textbf{H@10}\\
 \midrule
         \textbf{TransE} &  29.19 &  21.06  & 33.20&44.23&  30.73 &  23.45  & 35.18 &43.37\\
         \textbf{RotatE} &  33.67 &  26.80  & 36.68 &46.76&  34.95 &  29.10  & 38.35 &45.30\\
         \textbf{TuckER} &  30.39 &  24.44  & 32.91 &41.25&  37.05 &  34.59  & 38.43 &41.45\\
         \midrule
         \textbf{IKRL}&  32.36 &  26.11  & 34.75 &44.07&  33.22 &  30.37  & 34.28 &38.60\\
         \textbf{TransAE}&  30.00 &  21.23  & 34.91 &44.72  &  28.10 &  25.31  & 29.10&33.03\\
         \textbf{KG-Bert}& 28.68 & 21.12  & 32.57 &43.46&  - &  -  & - &-  \\
         \textbf{KGLM}& 34.12 & 27.01  & 36.87 &46.62&- & - & - &- \\
         \textbf{FLT-LM}&32.75  &25.89 & 32.87&44.56& - & -  & - &-  \\
 \textbf{OTKGE}& 34.36 & 28.85  & 34.36 &36.25& 35.51 & 31.97  & 37.18 &41.38\\
 \textbf{MMRNS}& 35.03& 28.59 & 37.49 &47.47& 35.93& 30.53 & 39.07 &45.47\\
 \textbf{VISTA}& 32.91 & 26.12  & 35.38 &45.61& 30.45 & 24.87  & 32.39 &41.53\\
 \textbf{MANS}& 30.88& 24.89 & 33.63 &41.78& 29.03& 25.25 & 31.35 &34.49\\
 \textbf{AdaMF}& 34.27& 27.21 & 37.86 &47.21& 38.06& 33.49 & \uline{40.44}&\uline{45.48}\\
 \textbf{MyGO}& 36.10 & 29.78  & 38.54&47.75& \uline{38.44} & 35.01  & 39.84 &44.19\\
 \textbf{MOMOK}& 35.89& \uline{30.38} & 37.54 &46.13& 37.91& \uline{35.09}  & 39.20 &43.20\\
 \midrule
\textbf{ KG-Llama-7b}& -& 20.20 & -&-& -& - & -&-\\
 \textbf{GPT 3.5 Turbo}& -& 22.66 & -&-& -& - & -&-\\
 \textbf{K-ON}& \uline{36.64}& 30.05& \uline{38.72}&\uline{48.26}& -& - & -&-\\
 \midrule
 \textbf{our model} & \textbf{43.27}& \textbf{36.02}& \textbf{46.51}&\textbf{57.12}& \textbf{43.62}& \textbf{37.68}& \textbf{45.78}&\textbf{54.66}\\ 
 Improvements & +18.1\% & +18.6\% & +20.1\% & +18.4\% & +13.5\% & +7.4\% & +13.2\% & +20.2\% \\
 \bottomrule
    \end{tabular}
\caption{Full results on the MKG-W and MKG-Y datasets.}
\label{tab:full-results}
\end{table*}

\subsection{Additional Results on WN18RR}
As our method involves registering all entities and relations into the LLM's vocabulary, we conducted additional experiments on WN18RR \cite{dettmers2018convolutional} to evaluate its performance on a dataset with a significantly larger entity space. While the datasets in our main settings contain at most 15,000 entities, WN18RR comprises 40,943 entities. We believe these supplementary experiments further demonstrate the generalizability and scalability of our approach.
% \subsection{Additional Results on WN18RR and Wikidata5M} 
% We present some more results on WN18RR and Wikidata5M, showing in the following Table \label{tab:additional} . The results indicate that KGT still performs well on classic KGC benchmarks, particularly in large-scale settings. For instance, Wikidata5M comprises over 4.8 million entities and 21 million triplets. The strong performance on such a large-scale benchmark highlights KGT's exceptional scalability and its potential for broad applicability in real-world, massive-scale knowledge representation scenarios.
\begin{table*}
    \centering
    \begin{tabular}{ c  cccc}
    \toprule
\multirow{2}{*}{\textbf{Methods}}  & \multicolumn{4}{c}{\textbf{WN18RR}} \\
  \cline{2-5}
 & \textbf{MRR$\uparrow$}& \textbf{Hits@1$\uparrow$}& \textbf{Hits@3$\uparrow$}&\textbf{Hits@10$\uparrow$}\\
  \midrule
  TransE& .232& .061& .366&.522\\
 RotatE& .476& .428& .492&.571\\
 TuckER& .470& .443& .526&.526\\
 \midrule
   KG-BERT& .216& .041& .302&.524\\
 StAR& .401& .243& .491&.709\\
 KGLM& .467& .330& .538&\uline{.741}\\
 \midrule
   GPT-3.5& -& .212& -&-\\
 Llama-2-13B& -& .315& -&-\\
 KICGPT& .549& .474& \uline{.585}&.641\\
 MKGL& \uline{.552}& \uline{.500}& .577&.656\\
   \midrule
    KGT& .622& .524& .679&.811\\
    \bottomrule
    \end{tabular}
\caption{Additional results on WN18RR. }
\label{tab:additional-datasets}
\end{table*}

% \subsection{ Results of different LLM}

% \begin{table}
% % \centering
% \begin{tabular}{c|cccc}
% \toprule
% \textbf{Model}&\textbf{MRR}&\textbf{Hits@1$\uparrow$}&\textbf{Hits@3$\uparrow$}&\textbf{Hits@10$\uparrow$}\\
% \midrule
%  Llama-2-7B& 43.62& 37.68& 45.78&54.66\\
%  Llama-3-1B& & & &\\
%  Llama-3-3B& 42.07& 36.05& 44.34&53.48\\
% Llama-3-8B&42.38&36.19&45.07&53.74\\
% \bottomrule
% \end{tabular}
% \caption{different model of MKG-Y.}
% \label{tab:diff-llm}
% \end{table}

% \subsection{Full results of ablation study}
% 开销对比表格
% \begin{table}[htbp]
% \centering
% \renewcommand{\arraystretch}{1.2}
% \setlength{\tabcolsep}{3pt}      

% \resizebox{\linewidth}{!}{
%     \begin{tabular}{c|cc|cc}
%     \toprule
%     % 表头第一行
%     \multirow{2}{*}{\textbf{Methods}} & \multicolumn{2}{c|}{\textbf{MKG-W}} & \multicolumn{2}{c}{\textbf{DB15K}} \\
%     % 局部横线
%     \cmidrule(lr){2-3} \cmidrule(lr){4-5} 
%     % 表头第二行
%      & \shortstack{\textbf{Trainable}\\\textbf{Params (M)}} & \textbf{H@1}& \shortstack{\textbf{Trainable}\\\textbf{Params (M)}} & \textbf{H@1}\\
%     \midrule
%     KG-Llama & 13& 20.20& 13& 33.55\\
%     MKGL     & 17 & 00.00 & 17 & 00.00 \\
%     K-ON     & 215 & 30.05& 208 & 30.13\\
%     KGT      & 19 & 38& 18 & 34\\
%     \bottomrule
%     \end{tabular}
% } % resizebox 的结束括号

% \caption{Computational cost of different LLM-based KGC methods.}
% \label{tab:efficiency}
% \end{table}
\end{document}